\documentclass[11pt,a4paper]{article}
\usepackage[hyperref]{acl2019}
\usepackage{times}
\usepackage{latexsym}
\usepackage{color}
\usepackage{times}
\usepackage{url}
\usepackage{latexsym}
\usepackage{xspace}
\usepackage{booktabs}
\usepackage{color}
\usepackage{graphicx}
\usepackage{tabulary}
\usepackage{enumitem}
\usepackage{url}
\usepackage{amsfonts}

\newenvironment{tightitemize}%
  {\begin{itemize}[topsep=0pt, partopsep=0pt] %
    \setlength{\itemsep}{0pt}%
    \setlength{\parskip}{0pt}%
    }%
  {\end{itemize}}
  \usepackage{booktabs}
\usepackage{subcaption}
\usepackage{lipsum}

  \usepackage{color}
  {\begin{enumerate}â \footnotesize%
    \setlength{\itemsep}{0pt}%
    \setlength{\parskip}{0pt}%
    }%
  {\end{enumerate}}


\usepackage{microtype}
\usepackage{multirow}
\usepackage{verbatim}
\usepackage{amsmath,amsthm,amssymb}
\usepackage{array}
\usepackage[scaled=0.86]{helvet}
\usepackage{ifthen}
\usepackage{courier}
\usepackage[boxed]{algorithm}
\usepackage{caption}
\usepackage{algpseudocode}

\belowdisplayskip \abovedisplayskip

\usepackage{amsmath}

\usepackage{url}

\aclfinalcopy 

\usepackage{CJKutf8}

\title{Is Word Segmentation Necessary for Deep Learning of Chinese Representations?}
\author{ Yuxian Meng$^{*1}$, Xiaoya Li$^{*1}$, Xiaofei Sun$^{1}$, Qinghong Han$^1$\\ {\bf Arianna Yuan$^{1,2}$, and Jiwei Li$^1$ }\\
~~\\$^1$ Shannon.AI
~~\\$^2$ Computer Science Department, Stanford University \\~~\\
\{ yuxian\_meng, xiaoya\_li, xiaofei\_sun, qinghong\_han \\ arianna\_yuan, jiwei\_li\}@shannonai.com
}

\date{}
\newcommand{\Word}{the word-based model\xspace}
\newcommand{\Char}{the char-based model\xspace}
\newcommand{\Words}{word-based models\xspace}
\newcommand{\Chars}{char-based models\xspace}
\newcommand{\sts}{\textsc{seq2seq}\xspace}

\begin{document}
\begin{CJK*}{UTF8}{gbsn}

\maketitle
\begin{abstract}
Segmenting a chunk of text into words is usually the first step of processing Chinese text, but its necessity has rarely been explored.  

In this paper, we ask the fundamental question of whether Chinese word segmentation (CWS) is necessary for deep learning-based Chinese Natural Language Processing. We benchmark neural word-based models which rely on word segmentation 
against neural char-based models which do not involve word segmentation in four end-to-end NLP benchmark tasks: language modeling, machine translation, sentence matching/paraphrase and text classification. Through direct  comparisons between these two types of models, we find that char-based models consistently outperform word-based models.
 
Based on these observations, we conduct comprehensive 
experiments to study why word-based models underperform char-based models in these deep learning-based NLP tasks.
We show that it is because word-based models  are more vulnerable to data sparsity and the presence of out-of-vocabulary (OOV) words, and thus more prone to overfitting. 
 We hope this paper could encourage researchers in the community to rethink the necessity of word segmentation in deep learning-based Chinese Natural Language Processing. \footnote{Yuxian Meng and Xiaoya Li contribute equally to this paper.}
\footnote{Paper to appear at ACL2019.}

\end{abstract}

\section{Introduction}
There is a key difference between English (or more broadly, languages that use some form of the Latin alphabet) and Chinese (or other languages that do not have obvious word delimiters such as Korean and Japanese) : words in English can be easily recognized 
since the space token is a good approximation of a word divider, whereas no word divider is present between words in written
Chinese sentences. 
This gives rise to the task of Chinese Word Segmentation (CWS) \cite{zhang2003hhmm,peng2004chinese,huang2007chinese,zhao2006improved,zheng2013deep,zhou2017word,yang2017neural,yang2018subword}.
In the context of deep learning, the segmented words are usually treated as the basic units for operations (we call these models the word-based models for the rest of this paper). 
  Each segmented word is  
 associated with a fixed-length vector representation, which will be processed by deep learning models in the same way as how English words are processed. 
Word-based models come with a few fundamental disadvantages, as will be discussed below. 

Firstly, word data sparsity inevitably leads to overfitting and the ubiquity of OOV words limits the model's learning capacity. Particularly, Zipf's law applies to most languages including Chinese. Frequencies of many Chinese  words are extremely small, making the model impossible to fully learn their semantics. 
Let us take the widely used Chinese Treebank dataset (CTB) as an example \cite{xia2000part}. Using Jieba,\footnote{\url{https://github.com/fxsjy/jieba}} the most widely-used
opensourced Chinese word segmentation system,  to segment the CTB, we end up with a dataset consisting of 615,194 words with 50,266 distinct words. 
Among the 50,266 distinct words,  24,458 words appear only once, amounting to 48.7\% of the total vocabulary, yet they only take up 4.0\% of the entire corpus. 
If we increase the frequency bar to 4,  we get 38,889 words appearing less or equal to 4 times, which contribute to 77.4\% of the total vocabulary but only 10.1\% of the entire corpus. Statistics are given in Table~\ref{tab:statistics of CTB by jieba}.
\begin{table}
\small
\begin{center}
\begin{tabular}{cccc}
bar& \# distinct  & prop of vocab &prop of corpus  \\\hline
$\infty$ & 50,266 & 100\% & 100\% \\
4 & 38,889 & 77.4\% & 10.1\% \\
1 & 24,458 & 48.7\% & 4.0\%\\\hline
\end{tabular}
\caption{Word statistics of Chinese TreeBank.}
\label{tab:statistics of CTB by jieba}
\end{center}
\end{table}
This shows that the word-based data is very sparse. The data sparsity issue is likely to induce overfitting, since more words means a larger number of parameters. In addition, since it is unrealistic to maintain a huge word-vector table, many words are treated as OOVs, which may further constrain the model's learning capability. 

Secondly, the state-of-the-art word segmentation performance is far from perfect, the errors of which would bias downstream NLP tasks. Particularly, CWS is a relatively hard and complicated task, primarily because word boundary of Chinese words is usually quite vague.
As discussed in \newcite{chen2017adversarial}, 
different linguistic perspectives have
 different criteria for CWS \cite{chen2017adversarial}. 
As shown in Table 1, 
in the two most widely adopted CWS datasets  PKU  \cite{yu2001processing} and CTB \cite{xia2000part}, the same sentence is segmented  differently. 

Thirdly, if we ask the fundamental problem of how much benefit word segmentation may provide, 
it is all about how much additional semantic information 
is present in a labeled CWS dataset. 
After all, the fundamental 
difference between \Words and \Chars is whether teaching signals from the CWS labeled dataset are utilized. 
Unfortunately, the answer to this question remains unclear. 
For example. in machine translation we usually have millions of training examples. 
The 
 labeled CWS dataset is relatively small (68k sentences for CTB and 21k for PKU),  
and 
the domain is relatively narrow. It is not clear 
that CWS dataset is sure to introduce a performance boost. 

\begin{table}
\small
\center
\begin{tabular}{|c|c|c|c|c|c|}\hline
Corpora&Yao&Ming&reaches&\multicolumn{2}{|c|}{the final} \\\hline
CTB& \multicolumn{2}{|c|}{姚明}& 进入&\multicolumn{2}{|c|}{总决赛}\\\hline
PKU&姚&明&进入&总&决赛 \\\hline
\end{tabular}
\caption{CTB and PKU have different segmentation criteria \cite{chen2017adversarial}. }
\end{table}

Before neural network models became popular, there were discussions on whether CWS is necessary and how much improvement it can bring about. 
 In information retrieval(IR), \newcite{foo2004chinese} discussed 
CWS's effect on IR systems and revealed that segmentation approach has an effect on IR effectiveness as long as the \textsc{same} segmentation method
is used for query and document, and that CWS does not always work better than models without segmentation.  
In cases where CWS does lead to better performance, the gap between \Words and \Chars can be closed if bigrams of characters are used in \Chars. 
In the phrase-based machine translation, 
\newcite{xu} reported that CWS only showed non-significant improvements over models without word segmentation. \newcite{zhao2013empirical} found that segmentation itself does not guarantee better MT performance and it is not key to MT improvement. 
For text classification, \newcite{liu2007chinese} compared a naïve character bigram model with word-based models, and concluded that CWS is not necessary for text classification. 
Outside the literature of computational linguistics, there have been discussions in the field of cognitive science.  
Based on eye movement data, \newcite{tsai2003chinese} found that fixations of Chinese readers do not land more frequently on the centers of Chinese words, suggesting that characters, rather than words, should be the basic units of Chinese reading comprehension. 
Consistent with this view, \newcite{bai2008reading} found that Chinese readers read unspaced text as fast as word spaced text. 

In this paper, we ask the fundamental question of whether word segmentation is necessary for deep learning-based Chinese natural language processing.
We first benchmark  word-based models against char-based models (those do not involve Chinese word segmentation). We run apples-to-apples comparison between these two types of models on four NLP tasks: 
 language modeling, document classification, machine translation and sentence matching. 
We  observe that char-based models consistently outperform word-based model.
We also compare char-based models with word-char hybrid models \cite{yin2016multi,dong2016character,yu2017joint}, and observe that 
\Chars perform better or at least as good as the hybrid model, indicating that \Chars already encode 
sufficient
semantic information. 

It is also crucial to understand the inadequacy of \Words. To this end, we perform comprehensive analyses on the behavior of word-based models and char-based models. We identify the major factor contributing to the disadvantage of \Words, i.e., data sparsity, which in turn leads to overfitting, prevelance of OOV words, and weak domain transfer ability.

Instead of making a conclusive (and arrogant) argument that Chinese word segmentation is not necessary, we hope this paper could 
foster more discussions and explorations on the necessity of the long-existing task of CWS in the community, alongside with its underlying mechanisms. 

\section{Related Work}
Since the First International Chinese Word Segmentation Bakeoff in 2003 \cite{sproat2003first} , a lot of effort has been made on Chinese word segmentation. 

Most of the models in the early years are based on a dictionary, which is pre-defined and thus independent of the Chinese text to be segmented. 
The simplest but 
remarkably robust 
model is the maximum matching model \cite{jurafsky2014speech}. 
The simplest version of it is 
the left-to-right maximum matching model ({\it maxmatch}). Starting with the beginning of a string, maxmatch chooses the longest word in the dictionary that matches the current position, and advances to the end of the matched word in the string. 
Different models are proposed based on different segmentation criteria \cite{huang2007chinese}.

With the rise of statistical machine learning methods, 
 the task of CWS is formalized as a tagging task, i.e., assigning a BEMS label to each character of a string that indicates whether the character is the start of a word(Begin), the end of a word(End), inside a word (Middel) or a single word(Single). 
 Traditional sequence labeling models such as HMM, MEMM and CRF are widely used \cite{lafferty2001conditional,peng2004chinese,zhao2006improved,carpenter2006character}. . 

Neural CWS  Models such as RNNs, LSTMs \cite{hochreiter1997long} and CNNs \cite{krizhevsky2012imagenet,kim2014convolutional} not only provide a more flexible way to incorporate 
context semantics into tagging models but also relieve researchers from the massive work of feature engineering. 
Neural models for the CWS task have become very popular these years \cite{chen2015long,chen2015gated,cai2016neural,yao2016bi,chen2016feature,zhang2016transition,chen2017adversarial,yang2017neural,cai2017fast,zhang2017segmenting}. Neural  representations can be  used either as a set of CRF features or as input to the decision layer.

\section{Experimental Results}
In this section, we evaluate the effect of word segmentation in deep learning-based Chinese NLP in  four tasks, language modeling, 
machine translation, text classification and sentence matching/paraphrase.
To enforce apples-to-apples comparison,
for both \Word and \Char, we use grid search to tune all important hyper-parameters such as learning rate, batch size, dropout rate, etc. 
 
\subsection{ Language Modeling}
We evaluate the two types of models on Chinese Tree-Bank 6.0 (CTB6). 
We followed the standard protocol, by which the dataset was
 split 
  into 80\%, 10\%, 10\%  for training, validation and test. 
  The task is formalized  as predicting the upcoming word given previous context representations. The text is segmented using Jieba.\footnote{\url{https://github.com/fxsjy/jieba}}
  An upcoming word is predicted given the previous context representation.
For different settings, 
context representations are obtained using \Char and \Word.
LSTMs are used to encode characters and words. 
\begin{table}
\small
\center
\begin{tabular}{ |ccc| }\hline
model &dimension & ppl  \\ \hline\hline
word  & 512& 199.9  \\\hline
char  &512& 193.0  \\\hline
word&2048&182.1  \\\hline
char&2048&170.9  \\\hline\hline
hybrid (word+char)& 1024+1024 & 175.7 \\\hline
hybrid (word+char)& 2048+1024 & 177.1 \\\hline
hybrid (word+char)& 2048+2048 & 176.2 \\\hline
hybrid (char only)& 2048 & 171.6 \\\hline
\end{tabular}
\caption{Language modeling perplexities in different models. }
\label{ppl}
\end{table}

Results are given in Table \ref{ppl}.  
In both settings, 
the char-based model significantly outperforms the word-based model.  
In addition to Jieba, we also used the Stanford CWS package \cite{monroe2014word} and the LTP package \cite{che2010ltp}, which resulted in similar findings. 

It is also interesting to see results from the 
hybrid model \cite{yin2016multi,dong2016character,yu2017joint}, which associates each word with  a representation and each char with a representation. 
A word representation is obtained by 
 combining the vector of its constituent word and vectors of the remaining 
characters. 
Since a Chinese word can contain an arbitrary number of characters, 
CNNs are applied to the combination of characters vectors \cite{kim2016character} to keep the dimensionality of the output representation invariant. 

We use {\it hybrid (word+char)} to denote the standard hybrid model that uses both char vectors and word vectors. 
For comparing purposes, we also implement a pseudo-hybrid model, denoted by {\it hybrid (char only)}, in which we do use a word segmentor to segment the texts, but word 
representations are obtained only using embeddings of their constituent characters.
We tune hyper-parameters such as vector dimensionality,  learning rate and batch size for all models.  

Results are given in Table~\ref{ppl}.  As can be seen, \Char not only outperforms \Word, but also the hybrid (word+char) model by a large margin. 
 The hybrid (word+char) model outperforms \Word. This means that  characters already encode all the semantic information needed and adding word embeddings would backfire. 
The hybrid (char only) model  performs similarly to \Char, suggesting that word segmentation does not provide any additional information. 
It outperforms \Word, which can be explained by that the hybrid (char only) model computes word representations 
only based on characters, and thus do not suffer from the data sparsity issue, OOV issue and the overfitting issue of \Word.

In conclusion, for the language modeling task on CTB, word segmentation does not provide any additional performance boost, and including
word embeddings worsen the result.

\begin{table*}[!ht]
\small
\center
\begin{tabular}{|c|ccc|cc|cc|}\hline
\multirow{ 2}{*}{TestSet} &\multirow{ 2}{*}{Mixed RNN}&\multirow{ 2}{*}{Bi-Tree-LSTM}&\multirow{ 2}{*}{PKI}& Seq2Seq& Seq2Seq & Seq2Seq (word)&Seq2Seq (char)  \\
&&& & +Attn (word) & +Attn (char)&+Attn+BOW  & +Attn+BOW \\\hline
MT-02 & 36.57&36.10 &  39.77&  35.67& 36.82 (+1.15)&37.70&40.14 (+0.37)\\
MT-03  &34.90&35.64 &33.64& 35.30 &36.27 (+0.97) &38.91 &40.29 (+1.38)\\
MT-04  & 38.60&36.63&  36.48& 37.23 &37.93 (+0.70)&40.02& 40.45 (+0.43) \\
MT-05 & 35.50 & 34.35&33.08  & 33.54 &34.69 (+1.15)&36.82&36.96 (+0.14) \\
MT-06   & 35.60&30.57&  32.90& 35.04 &35.22 (+0.18)&35.93& 36.79 (+0.86) \\
MT-08  & --& --&24.63&26.89  &  27.27 (+0.38) &27.61& 28.23 (+0.62)\\
Average &--&--&32.51& 33.94 &34.77 (+0.83) &36.51 & 37.14 (+0.63)\\\hline
\end{tabular}
\caption{Results of different models on the Ch-En machine translation task. Results of Mixed RNN \cite{li2017modeling}, Bi-Tree-LSTM \cite{chen2017improved} and PKI \cite{zhang2018prior}  are copied from the original papers.}
\label{MT}
\end{table*}
\begin{table*}[!ht]
\small
\center
\begin{tabular}{|c|cc|cc|}\hline
\multirow{ 2}{*}{TestSet} & Seq2Seq& Seq2Seq & Seq2Seq &Seq2Seq (char)  \\
 & +Attn (word) & +Attn (char)&+Attn+BOW  & +Attn+BOW \\\hline
MT-02 &  42.57& 44.09 (+1.52)&43.42&46.78 (+3.36)\\
MT-03  & 40.88 &44.57 (+3.69) &43.92 &47.44 (+3.52)\\
MT-04  & 40.98 &44.73 (+3.75)&43.35& 47.29 (+3.94) \\
MT-05  & 40.87 &42.50 (+1.63)&42.63&44.73 (+2.10) \\
MT-06   & 39.33 &42.88 (+3.55)&43.31& 46.66 (+3.35) \\
MT-08  &33.52  &  35.36 (+1.84) &35.65& 38.12 (+2.47)\\
Average & 39.69 &42.36 (+2.67) &42.04 & 45.17 (+3.13)\\\hline\hline
\end{tabular}
\caption{Results on the En-Ch machine translation task.}
\label{MT-ch}
\end{table*}

\subsection{Machine Translation}
In our experiments on machine translation, we use the standard Ch-En setting.
The training set consists of  1.25M sentence pairs extracted from the LDC corpora.\footnote{LDC2002E18, LDC2003E07,
LDC2003E14, Hansards portion of LDC2004T07,
LDC2004T08 and LDC2005T06.} 
The validation set is from NIST 2002 and the models are evaluated on NIST 2003, 2004, 2005, 2006 and 2008.
We followed exactly the common setup in \newcite{ma2018bag,chen2017improved,li2017modeling,zhang2018prior},
which use top 30,000 English words and 27,500 Chinese words.
For \Char, vocab size is set to 4,500.  
We report results in both the Ch-En and the En-Ch settings. 

Regarding the implementation, we  compare \Chars with \Words
under the standard framework of \sts+attention   \cite{sutskever2014sequence,luong2015effective}.
The current state-of-the-art model is from \newcite{ma2018bag}, which
uses both the sentences (seq2seq) and the bag-of-words as targets in
the training stage. We simply change the word-level encoding in \newcite{ma2018bag} to char-level encoding. 
For En-Ch translation,  we use the same dataset to train and test both models. 
As in \newcite{ma2018bag}, the dimensionality for word vectors and char vectors is set to 512.\footnote{We found that transformers \cite{vaswani2017attention} underperform  LSTMs+attention on this dataset. We conjecture that this is due to the relatively small size (1.25M) of the training set. The size of the dataset in \newcite{vaswani2017attention} is at least 4.5M. LSTMs+attention is usually more robust on smaller datasets, due to the smaller number of parameters. }

Results for Ch-En are shown in Table \ref{MT}. As can be seen, for the vanilla \sts+attention model, \Char outperforms \Word across all datasets, yielding an average performance boost of +0.83. 
The same pattern applies to the bag-of-words framework in \newcite{ma2018bag}. When changing \Word to \Char, we are able to 
obtain a performance boost of +0.63. As far as we are concerned, 
this is the best result on this 1.25M Ch-En dataset. 
\begin{table*}
   \small
   \center
    \begin{tabular}{cccccc}
        \toprule
        {\bf Dataset} & {\bf description} &{\bf char valid} & {\bf word valid} & {\bf char test} & {\bf word test} \\
        \midrule
        {\bf LCQMC} & 238.7K/8.8K/12.5K & 84.70 & 83.48 & {\bf 84.43}  (+1.34)& 83.09 \\
        {\bf BQ} & 100K/10K/10K & 82.59 & 79.63 & {\bf 82.19} (+2.90) & 79.29 \\
        \bottomrule
    \end{tabular}
    \caption{Results on the LCQMC and BQ corpus.}
    \label{bq}
\end{table*}
\begin{table*}[thb]
\centering
\small
\begin{tabular}{ccccll}
    \toprule
    {\bf Dataset} & {\bf description} & {\bf char valid} & {\bf word valid} & {\bf char test} & {\bf word test} \\
    \midrule 
    {\bf chinanews} & 1260K/140K/112K & 91.81 & 91.82 & 91.80 & {\bf 91.85} (+0.05)\\
    {\bf dianping} & 1800K/200K/500K & 78.80 & 78.47 & {\bf 78.76} (+0.36) & 78.40 \\
    {\bf ifeng} & 720K/80K/50K & 86.04 & 84.89 & {\bf 85.95} (+1.09) & 84.86 \\
    {\bf jd\_binary} & 3600K/400K/360K & 92.07 & 91.82 & {\bf 92.05} (+0.16) & 91.89 \\
    {\bf jd\_full} & 2700K/300K/250K & 54.29 & 53.60 & {\bf 54.18} (+0.81) & 53.37 \\
    \bottomrule
\end{tabular}
\caption{Results on the validation and the test set for text classification.}
\label{text-cls}
\end{table*}

Results for En-Ch are presented in Table \ref{MT-ch}. As can be seen, \Char outperforms \Word by a huge margin (+3.13), and this margin is greater than the improvement in the Ch-En translation task. 
This is because in Ch-En translation, the difference between word-based and char-based models is only present in the source encoding stage, whereas in  En-Ch translation it is present in both the source encoding and the target decoding stage. 
Another major reason that contributes to the inferior  performance of \Word is the UNK word at decoding time,
We also implemented the BPE subword model \cite{sennrich2015neural,sennrich2016edinburgh} on the Chinese target side.
The BPE model achieves a performance of 41.44 for the Seq2Seq+attn setting and 44.35 for bag-of-words, 
significantly outperforming \Word, but still underperforming \Char by about 0.8-0.9 in BLEU.

We conclude that for Chinese, generating characters has the advantage over generating words in deep learning decoding.

\subsection{Sentence Matching/Paraphrase}
There are two Chinese datasets similar to  the Stanford Natural Language Inference (SNLI) Corpus \cite{bowman2015large}: BQ and LCQMC, in which we need to assign a label to a pair of sentences depending on whether they share similar meanings.  
For the BQ dataset \cite{chen2018bq}, it contains 120,000 Chinese sentence pairs, and each pair is associated with a label indicating whether 
the two sentences 
 are of equivalent semantic meanings. 
The dataset is deliberately constructed so that sentences in some pairs may have significant word overlap  but complete different meanings, while others are the other way around.
For LCQMC \cite{liu2018lcqmc}, 
it
 aims at identifying whether two sentences have the same intention. 
This task is similar to but not exactly the same as the paraphrase detection task in BQ: two sentences can have different meanings but share the same intention. For example, the meanings of "{\it My phone is lost}" and "{\it I need a new phone}"  are different, but their intentions are the same: buying a new phone. 

Each pair of sentences in the BQ and the LCQMC dataset is associated with a binary label indicating whether the two sentences share the same intention, and the task can be formalized as predicting this binary label. 
To predict correct labels, a model needs to handle the semantics of the sub-units of a sentence, which makes the task very appropriate for examining 
the capability of semantic models. 

We compare  \Chars with \Words.  
For the \Words, texts are segmented using Jieba. 
The SOTA results on these two datasets is achieved by the bilateral
multi-perspective matching model (BiMPM) \cite{wang2017bilateral}.
We use the standard settings proposed by BiMPM, i.e. 200d word/char embeddings, which are randomly initialized.  

Results are shown in Table \ref{bq}. As can be seen, \Char significantly outperforms \Word by a huge margin, +1.34 on the LCQMC dataset and +2.90 on the BQ set. 
For this paraphrase detection task, the model needs to handle the interactions between sub-units of a sentence. 
We conclude that \Char is significantly better in this respect.

\subsection{Text Classification}
For text classification, we use the currently widely used benchmarks including:
\begin{tightitemize}
\item ChinaNews: Chinese news articles  split into 7 news categories. 
\item Ifeng: First paragraphs of Chinese news articles from 2006-2016. The dataset consists of 5 news
categories; 
\item JD\_Full:  product reviews in Chinese crawled from JD.com. The reviews are used to predict customers' ratings (1 to 5 stars), making the task a five-class classification problem.
\item JD\_binary: the same product reviews from JD.com.  We label 1, 2-star reviews as ``negative reviews''  and 4 and 5-star reviews as ``positive reviews'' (3-star reviews are ignored), making the task a binary-classification problem. 
 \item Dianping: 
Chinese restaurant reviews crawled from the online  review website 
Dazhong Dianping (similar to Yelp). We collapse the 1, 2 and 3-star reviews to ``negative reviews'' and 4 and 5-star reviews to ``positive reviews''.

 \end{tightitemize}
The datasets were first 
introduced in \newcite{zhang2017encoding}.
We trained the word-based version and the char-based version of bi-directional LSTM models  to solve this task. 
Results are shown in Table \ref{text-cls}.
As can be seen, the only dataset that \Char underperforms \Word is the chinanews dataset, but the difference is quite small (0.05). On all the other datasets, \Char significantly outperforms \Word.

\paragraph{Domain Adaptation Ability} \cite{daume2009frustratingly,jiang2008domain,zhuang2010cross}
refers to the ability of extending  a  model learned from one data distribution (the source domain) for a different (but related) data distribution (the target domain). 
Because of the data sparsity issue, we hypothesize that \Chars have greater domain adaptation ability than 
\Words. 

We test our hypothesis on different sentiment analysis datasets.
We  train \Word and \Char   on Dianping (2M restaurant reviews) and test the two models on jd\_binary (0.25M product reviews), as denoted by train\_dianping\_test\_jd. We also
train models on jd\_binary and test them on Dianping, as denoted by train\_jd\_test\_dianping). 
Results are given in Table~\ref{tab:Domain adaptation of Word and Char}. 
\begin{table}
\center
\small
\begin{tabular}{|c|c|c|}\hline
\multicolumn{3}{|c|}{train\_dianping\_test\_jd}   \\\hline
model & acc & proportion of sen  \\
&  &  containing OOV  \\\hline
word-based & 81.28\% & 11.79\% \\
char-based & 83.33\% & 0.56\% \\\hline\hline
\multicolumn{3}{|c|}{train\_jd\_test\_dianping}   \\\hline
model & acc & proportion of sen  \\
&  &  containing OOV  \\\hline
word-based & 67.32\% & 7.10\% \\
char-based & 67.93\% & 46.85\% \\\hline
\end{tabular}
\caption{Domain adaptation of \Word and \Char}
\label{tab:Domain adaptation of Word and Char}
\end{table}
As expected, \Char has more domain adaptation ability
and performs better 
 than \Word on both settings.  
The OOV issue is especially serious for \Word. In the train\_dianping\_test\_jd setting, 11.79\%
of the test sentences contain OOVs for \Word, whereas this number is only 0.56\% for \Char. Similar observation holds for 
the train\_jd\_test\_dianping setting.

\section{Analysis}
In this section, we aim at understanding why \Words underperform \Chars. We acknowledge that it is impossible to thoroughly inspect the inner mechanism of \Words, but we try our best to identify major factors explaining the inferiority of \Words.  
\begin{figure*}[!ht]
\center
\includegraphics[height=4.8cm,width=14.5cm]{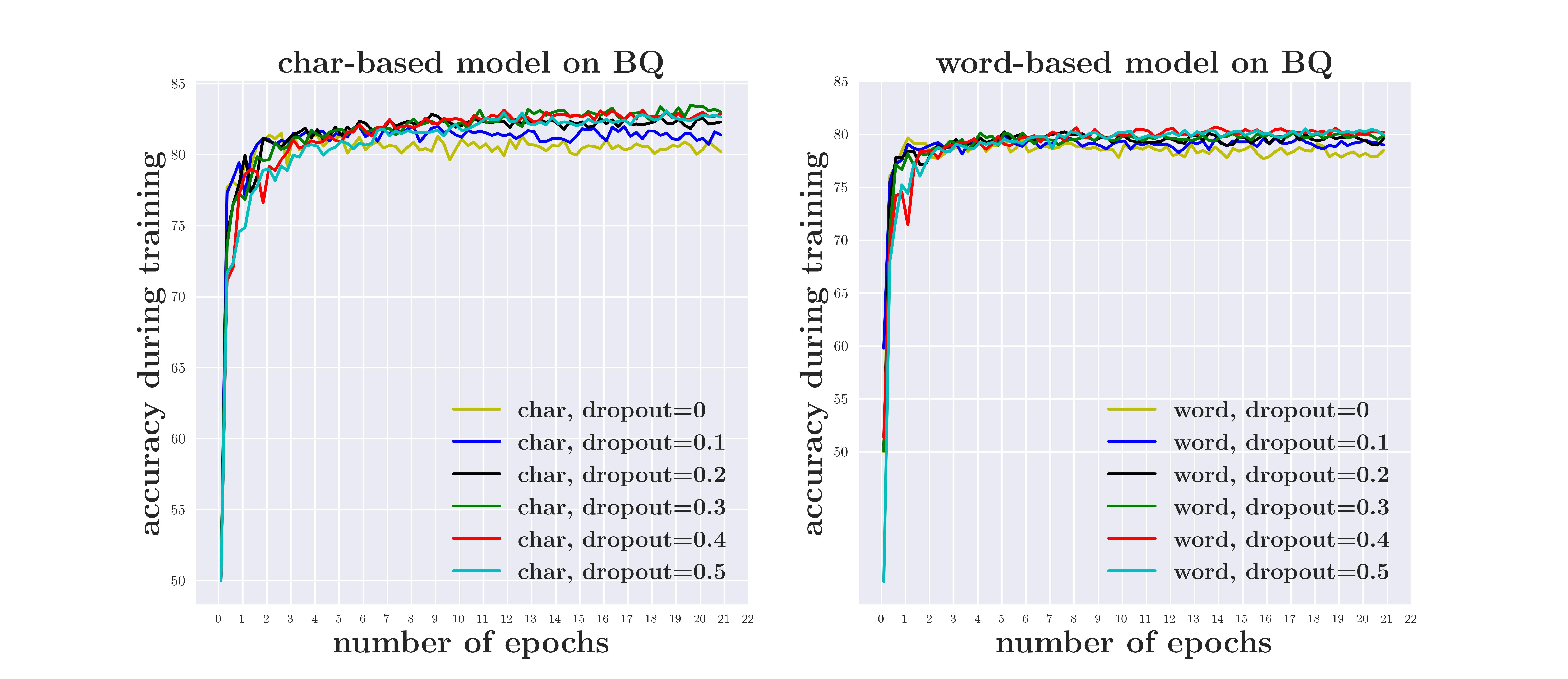}
\caption{Effects of dropout rates on \Char and \Word. }
\label{dr}
\end{figure*}
\begin{figure*}[!ht]
\center
\includegraphics[height=4.8cm,width=14.5cm]{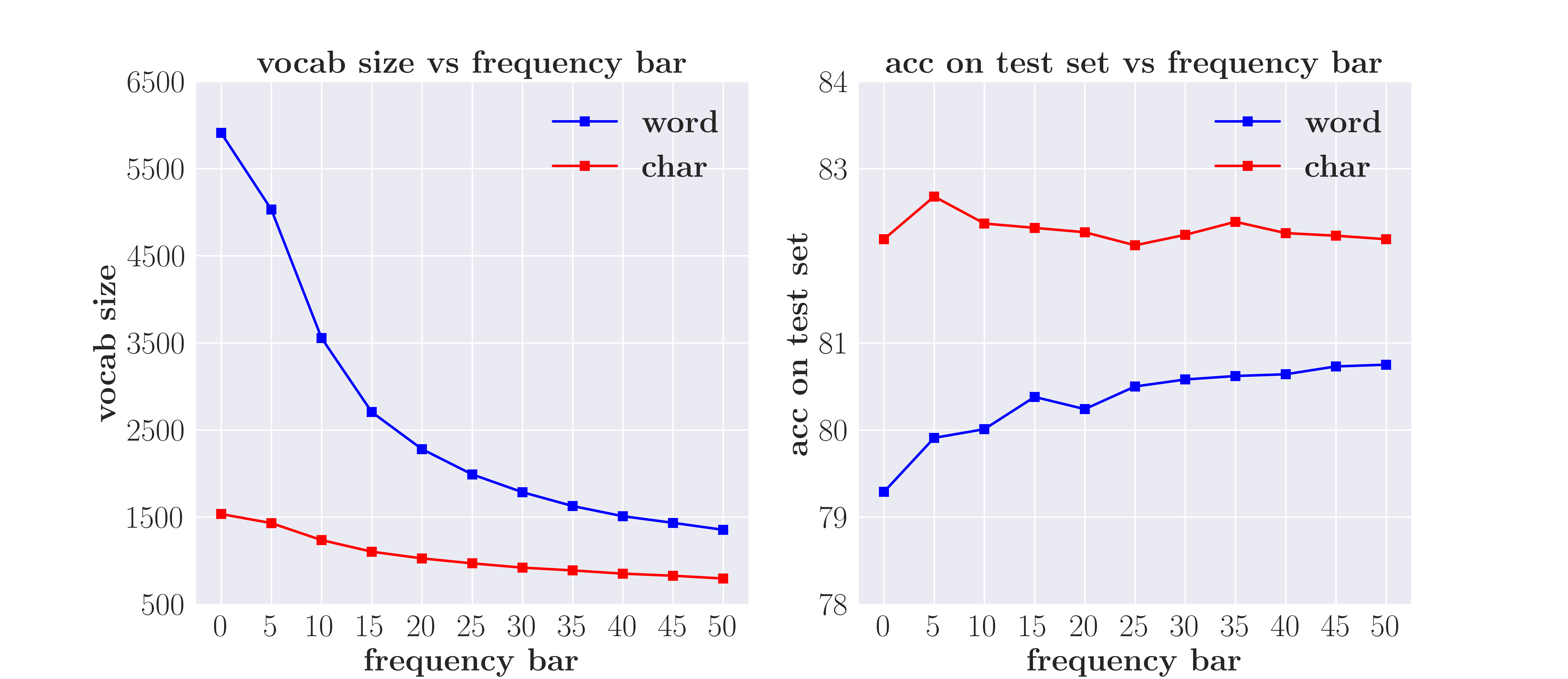}
\caption{Effects of data sparsity on \Char and \Word. }
\label{sparsity}
\end{figure*}
\begin{figure*}[!ht]
\center
\includegraphics[width=12cm]{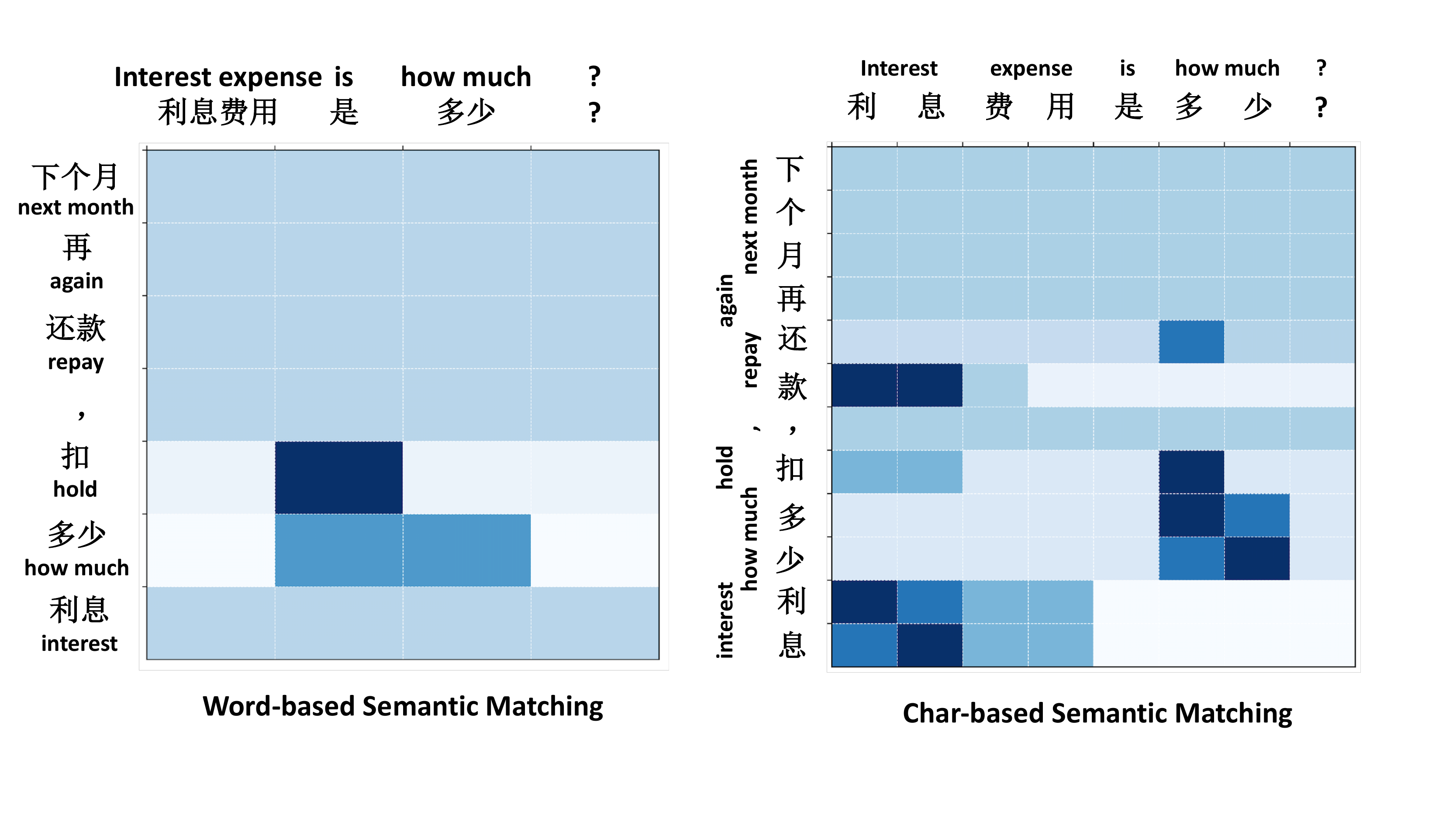}
\caption{Semantic matching between two Chinese sentences with char-based models and word-based models. }
\label{vis}
 \end{figure*}

\begin{figure*}[!ht]
\center
\includegraphics[height=4.8cm,width=14.5cm]{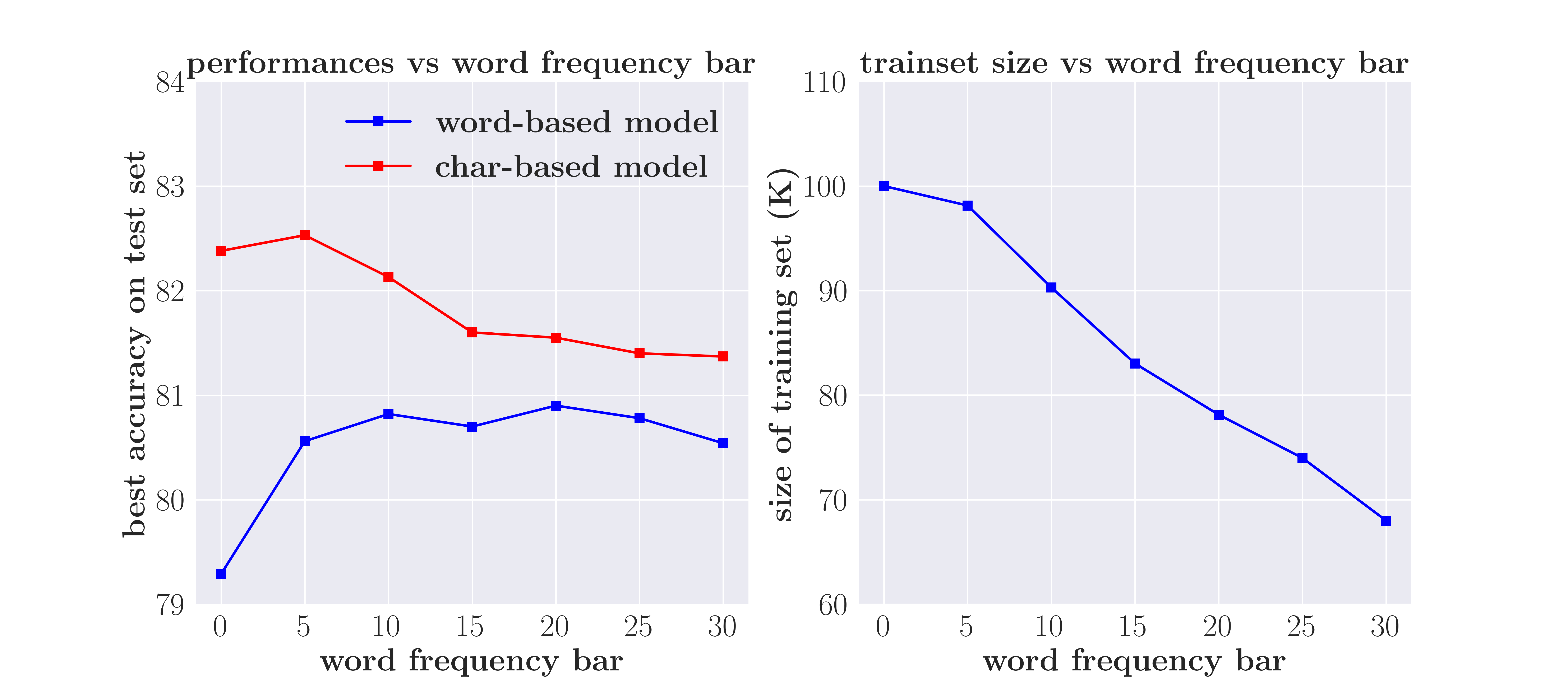}
\caption{Effects of removing training instances containing OOV words. }
\label{OOV}
\end{figure*}

\subsection{Data Sparsity}
A common method to avoid  vocabulary size getting too big is to set a frequency threshold, and use a special UNK token to denote all words whose frequency is below the threshold. 
The value of the frequency threshold is closely related to the vocabulary size, and consequently the number of parameters. 
Figure \ref{sparsity} shows the correlation between the vocabulary size and the frequency threshold, along with
 the correlation 
between model performances and the frequency threshold.
For both \Char and \Word, using all words/chars (threshold set to 0) leads to bad results. 
The explanation is intuitive: it is hard to learn the semantics of infrequent words/characters. 

For \Char, the best performance is obtained when character frequency threshold is set to 5, resulting in a vocabulary size of 1,432 and a medium character frequency of 72.
 For \Word, the best performance is obtained when word frequency threshold is set to 50, in which case the vocabulary size is 1,355 and the medium word frequency is 83. 
 As can be seen, the vocabulary size and the medium word frequency for the best word-based model is similar to those of the best char-based model. 
This means, for a given dataset, in order to learn the word/char semantics well, the model needs to have enough exposure to each word/character, the amount of which is approximately the same across different models. 
For \Word, this requirement is particularly hard to meet due to its sparseness. 

\subsection{Out-of-Vocabulary Words}
One possible explanation for the inferiority of \Word is that it contains too many OOVs. If so, we should be able to narrow or even close the gap between \Words and \Chars by decreasing the number of OOVs.
As discussed in Section 4.2, 
setting the frequency threshold low  to avoid OOVs 
will hinder the performance because it worsen the data sparsity issue. We thus use an alternative strategy: for different word-frequency thresholds, we remove sentences that contain word OOVs from all of the training, validation and test sets.  
Figure \ref{OOV} shows vocabulary sizes of the training set and accuracies   plotted against word frequency threshold.  
 As can be seen, the gap between the two types of models is gradually narrowed as we increase the word-frequency threshold. 
It is also interesting that the curve for \Char goes up slightly at the beginning and then goes down steadily. It is because the OOV issue is not severe for \Char and thus does not affect the performance much. However, as we remove more and more training examples, the shrinking training dataset creates a bigger problem.  
By contrast, for \Word, the performance keeps increasing even when the frequency threshold is set to 50, meaning that the positive influence of removing some OOVs outweighs the negative influence of eliminating some training data. 
In conclusion, \Word suffers from the OOV issue. This issue can be alleviated by reducing the number of OOVs in the datasets. 
\subsection{Overfitting}
The data sparsity issue leads to the fact that
word-based models have more parameters to learn, and thus are more prone to overfitting.  
We conducted experiments on the BQ dataset \cite{chen2018bq} and the results validate this point (Figure~\ref{dr}). To achieve the best results, a  larger dropout rate is needed for \Word (0.5) than \Char (0.3). This means overfitting is a severe issue for \Word. 
We also observe that curves with different dropout rates are closer together in \Words than in \Chars, which means the dropout technique is not enough to resolve the overfitting issue. \Char without dropout
 already achieves better performance (80.82) than \Word with the optimal dropout rate (80.65).

\subsection{Visualization}
The BQ semantic matching task aims at deciding whether two sentences have the same intention.
Figure \ref{vis} tangibly shows why \Char outperforms \Word. 
The heatmap denotes the attention matching values between tokens of two two sentences, 
computed by the BiPMP model \cite{wang2017bilateral}.  
The input two sentences are: (1) 利息费用是多少~ (how much is the interest expense), with segmented text being 利息费用~ (interest expense) 是~ (is) 多少~ (how much)  and (2) 下一个月还款要扣多少利息~ (how much interest do I have to pay if I repay the bill next month), with segmented text being 
下个月 (next month) 还款~ (repay), 扣~ (hold) 多少~ (how much) 利息 ~(interest). 
For word-based semantic matching, since 利息费用 (interest expense) is treated as a single word, 
it fails to be mapped to 利息 (interest). This is not the case with \Char since the same character in the two sentences
 are more easily mapped. 

\section{Conclusion}
In this paper, we ask the fundamental question of whether word segmentation is necessary for deep learning
of Chinese representations.
We benchmark such word-based models against char-based models  in four end-to-end NLP tasks,
 and enforce
apples-to-apples comparisons as much as possible. 
We  observe that char-based models consistently outperform word-based models.
Building upon these findings, we show that 
word-based models' inferiority is due to 
 the sparseness of word distributions, which leads to more out-of-vocabulary words, overfitting and lack of domain generalization ability.
We hope this paper will 
foster more discussions on the necessity of the long-existing task of CWS in the community. 
\end{CJK*}

\bibliography{acl2019}

\begin{thebibliography}{55}
\expandafter\ifx\csname natexlab\endcsname\relax\def\natexlab#1{#1}\fi

\bibitem[{Bai et~al.(2008)Bai, Yan, Liversedge, Zang, and
  Rayner}]{bai2008reading}
Xuejun Bai, Guoli Yan, Simon~P Liversedge, Chuanli Zang, and Keith Rayner.
  2008.
\newblock Reading spaced and unspaced chinese text: Evidence from eye
  movements.
\newblock \emph{Journal of Experimental Psychology: Human Perception and
  Performance}, 34(5):1277.

\bibitem[{Bowman et~al.(2015)Bowman, Angeli, Potts, and
  Manning}]{bowman2015large}
Samuel~R Bowman, Gabor Angeli, Christopher Potts, and Christopher~D Manning.
  2015.
\newblock A large annotated corpus for learning natural language inference.
\newblock \emph{EMNLP}.

\bibitem[{Cai and Zhao(2016)}]{cai2016neural}
Deng Cai and Hai Zhao. 2016.
\newblock Neural word segmentation learning for chinese.
\newblock \emph{ACL}.

\bibitem[{Cai et~al.(2017)Cai, Zhao, Zhang, Xin, Wu, and Huang}]{cai2017fast}
Deng Cai, Hai Zhao, Zhisong Zhang, Yuan Xin, Yongjian Wu, and Feiyue Huang.
  2017.
\newblock Fast and accurate neural word segmentation for chinese.
\newblock \emph{ACL}.

\bibitem[{Carpenter(2006)}]{carpenter2006character}
Bob Carpenter. 2006.
\newblock Character language models for chinese word segmentation and named
  entity recognition.
\newblock In \emph{Proceedings of the Fifth SIGHAN Workshop on Chinese Language
  Processing}, pages 169--172.

\bibitem[{Che et~al.(2010)Che, Li, and Liu}]{che2010ltp}
Wanxiang Che, Zhenghua Li, and Ting Liu. 2010.
\newblock Ltp: A chinese language technology platform.
\newblock In \emph{Proceedings of the 23rd International Conference on
  Computational Linguistics: Demonstrations}, pages 13--16. Association for
  Computational Linguistics.

\bibitem[{Chen et~al.(2017{\natexlab{a}})Chen, Huang, Chiang, and
  Chen}]{chen2017improved}
Huadong Chen, Shujian Huang, David Chiang, and Jiajun Chen. 2017{\natexlab{a}}.
\newblock Improved neural machine translation with a syntax-aware encoder and
  decoder.
\newblock \emph{ACL}.

\bibitem[{Chen et~al.(2018)Chen, Chen, Liu, Yang, Lu, and Tang}]{chen2018bq}
Jing Chen, Qingcai Chen, Xin Liu, Haijun Yang, Daohe Lu, and Buzhou Tang. 2018.
\newblock The bq corpus: A large-scale domain-specific chinese corpus for
  sentence semantic equivalence identification.
\newblock In \emph{Proceedings of the 2018 Conference on Empirical Methods in
  Natural Language Processing}, pages 4946--4951.

\bibitem[{Chen et~al.(2017{\natexlab{b}})Chen, Qiu, and
  Huang}]{chen2016feature}
Xinchi Chen, Xipeng Qiu, and Xuanjing Huang. 2017{\natexlab{b}}.
\newblock A feature-enriched neural model for joint chinese word segmentation
  and part-of-speech tagging.
\newblock \emph{IJCAI}.

\bibitem[{Chen et~al.(2015{\natexlab{a}})Chen, Qiu, Zhu, and
  Huang}]{chen2015gated}
Xinchi Chen, Xipeng Qiu, Chenxi Zhu, and Xuanjing Huang. 2015{\natexlab{a}}.
\newblock Gated recursive neural network for chinese word segmentation.
\newblock In \emph{Proceedings of the 53rd Annual Meeting of the Association
  for Computational Linguistics and the 7th International Joint Conference on
  Natural Language Processing (Volume 1: Long Papers)}, volume~1, pages
  1744--1753.

\bibitem[{Chen et~al.(2015{\natexlab{b}})Chen, Qiu, Zhu, Liu, and
  Huang}]{chen2015long}
Xinchi Chen, Xipeng Qiu, Chenxi Zhu, Pengfei Liu, and Xuanjing Huang.
  2015{\natexlab{b}}.
\newblock Long short-term memory neural networks for chinese word segmentation.
\newblock In \emph{Proceedings of the 2015 Conference on Empirical Methods in
  Natural Language Processing}, pages 1197--1206.

\bibitem[{Chen et~al.(2017{\natexlab{c}})Chen, Shi, Qiu, and
  Huang}]{chen2017adversarial}
Xinchi Chen, Zhan Shi, Xipeng Qiu, and Xuanjing Huang. 2017{\natexlab{c}}.
\newblock Adversarial multi-criteria learning for chinese word segmentation.
\newblock \emph{ACL}.

\bibitem[{Daum{\'e}~III(2007)}]{daume2009frustratingly}
Hal Daum{\'e}~III. 2007.
\newblock Frustratingly easy domain adaptation.
\newblock \emph{ACL}.

\bibitem[{Dong et~al.(2016)Dong, Zhang, Zong, Hattori, and
  Di}]{dong2016character}
Chuanhai Dong, Jiajun Zhang, Chengqing Zong, Masanori Hattori, and Hui Di.
  2016.
\newblock Character-based lstm-crf with radical-level features for chinese
  named entity recognition.
\newblock In \emph{Natural Language Understanding and Intelligent
  Applications}, pages 239--250. Springer.

\bibitem[{Foo and Li(2004)}]{foo2004chinese}
Schubert Foo and Hui Li. 2004.
\newblock Chinese word segmentation and its effect on information retrieval.
\newblock \emph{Information processing \& management}, 40(1):161--190.

\bibitem[{Hochreiter and Schmidhuber(1997)}]{hochreiter1997long}
Sepp Hochreiter and J{\"u}rgen Schmidhuber. 1997.
\newblock Long short-term memory.
\newblock \emph{Neural computation}, 9(8):1735--1780.

\bibitem[{Huang and Zhao(2007)}]{huang2007chinese}
Changning Huang and Hai Zhao. 2007.
\newblock Chinese word segmentation: A decade review.
\newblock \emph{Journal of Chinese Information Processing}, 21(3):8--20.

\bibitem[{Jiang(2008)}]{jiang2008domain}
Jing Jiang. 2008.
\newblock Domain adaptation in natural language processing.
\newblock Technical report.

\bibitem[{Jurafsky and Martin(2014)}]{jurafsky2014speech}
Dan Jurafsky and James~H Martin. 2014.
\newblock \emph{Speech and language processing}, volume~3.
\newblock Pearson London.

\bibitem[{Kim(2014)}]{kim2014convolutional}
Yoon Kim. 2014.
\newblock Convolutional neural networks for sentence classification.
\newblock \emph{EMNLP}.

\bibitem[{Kim et~al.(2016)Kim, Jernite, Sontag, and Rush}]{kim2016character}
Yoon Kim, Yacine Jernite, David Sontag, and Alexander~M Rush. 2016.
\newblock Character-aware neural language models.
\newblock In \emph{AAAI}, pages 2741--2749.

\bibitem[{Krizhevsky et~al.(2012)Krizhevsky, Sutskever, and
  Hinton}]{krizhevsky2012imagenet}
Alex Krizhevsky, Ilya Sutskever, and Geoffrey~E Hinton. 2012.
\newblock Imagenet classification with deep convolutional neural networks.
\newblock In \emph{Advances in neural information processing systems}, pages
  1097--1105.

\bibitem[{Lafferty et~al.(2001)Lafferty, McCallum, and
  Pereira}]{lafferty2001conditional}
John Lafferty, Andrew McCallum, and Fernando~CN Pereira. 2001.
\newblock Conditional random fields: Probabilistic models for segmenting and
  labeling sequence data.

\bibitem[{Li et~al.(2017)Li, Xiong, Tu, Zhu, Zhang, and Zhou}]{li2017modeling}
Junhui Li, Deyi Xiong, Zhaopeng Tu, Muhua Zhu, Min Zhang, and Guodong Zhou.
  2017.
\newblock Modeling source syntax for neural machine translation.
\newblock \emph{arXiv preprint arXiv:1705.01020}.

\bibitem[{Liu et~al.(2007)Liu, Allison, Guthrie, and Guthrie}]{liu2007chinese}
Wei Liu, Ben Allison, David Guthrie, and Louise Guthrie. 2007.
\newblock Chinese text classification without automatic word segmentation.
\newblock In \emph{Sixth International Conference on Advanced Language
  Processing and Web Information Technology (ALPIT 2007)}, pages 45--50. IEEE.

\bibitem[{Liu et~al.(2018)Liu, Chen, Deng, Zeng, Chen, Li, and
  Tang}]{liu2018lcqmc}
Xin Liu, Qingcai Chen, Chong Deng, Huajun Zeng, Jing Chen, Dongfang Li, and
  Buzhou Tang. 2018.
\newblock Lcqmc: A large-scale chinese question matching corpus.
\newblock In \emph{Proceedings of the 27th International Conference on
  Computational Linguistics}, pages 1952--1962.

\bibitem[{Luong et~al.(2015)Luong, Pham, and Manning}]{luong2015effective}
Minh-Thang Luong, Hieu Pham, and Christopher~D Manning. 2015.
\newblock Effective approaches to attention-based neural machine translation.
\newblock \emph{ACL}.

\bibitem[{Ma et~al.(2018)Ma, Sun, Wang, and Lin}]{ma2018bag}
Shuming Ma, Xu~Sun, Yizhong Wang, and Junyang Lin. 2018.
\newblock Bag-of-words as target for neural machine translation.
\newblock \emph{arXiv preprint arXiv:1805.04871}.

\bibitem[{Monroe et~al.(2014)Monroe, Green, and Manning}]{monroe2014word}
Will Monroe, Spence Green, and Christopher~D Manning. 2014.
\newblock Word segmentation of informal arabic with domain adaptation.
\newblock In \emph{Proceedings of the 52nd Annual Meeting of the Association
  for Computational Linguistics (Volume 2: Short Papers)}, volume~2, pages
  206--211.

\bibitem[{Peng et~al.(2004)Peng, Feng, and McCallum}]{peng2004chinese}
Fuchun Peng, Fangfang Feng, and Andrew McCallum. 2004.
\newblock Chinese segmentation and new word detection using conditional random
  fields.
\newblock In \emph{Proceedings of the 20th international conference on
  Computational Linguistics}, page 562. Association for Computational
  Linguistics.

\bibitem[{Sennrich et~al.(2016{\natexlab{a}})Sennrich, Haddow, and
  Birch}]{sennrich2016edinburgh}
Rico Sennrich, Barry Haddow, and Alexandra Birch. 2016{\natexlab{a}}.
\newblock Edinburgh neural machine translation systems for wmt 16.
\newblock \emph{WMT}.

\bibitem[{Sennrich et~al.(2016{\natexlab{b}})Sennrich, Haddow, and
  Birch}]{sennrich2015neural}
Rico Sennrich, Barry Haddow, and Alexandra Birch. 2016{\natexlab{b}}.
\newblock Neural machine translation of rare words with subword units.
\newblock \emph{ACL}.

\bibitem[{Sproat and Emerson(2003)}]{sproat2003first}
Richard Sproat and Thomas Emerson. 2003.
\newblock The first international chinese word segmentation bakeoff.
\newblock In \emph{Proceedings of the second SIGHAN workshop on Chinese
  language processing-Volume 17}, pages 133--143. Association for Computational
  Linguistics.

\bibitem[{Sutskever et~al.(2014)Sutskever, Vinyals, and
  Le}]{sutskever2014sequence}
Ilya Sutskever, Oriol Vinyals, and Quoc~V Le. 2014.
\newblock Sequence to sequence learning with neural networks.
\newblock In \emph{Advances in neural information processing systems}, pages
  3104--3112.

\bibitem[{Tsai and McConkie(2003)}]{tsai2003chinese}
Jie-Li Tsai and George~W McConkie. 2003.
\newblock Where do chinese readers send their eyes?
\newblock In \emph{The Mind's Eye}, pages 159--176. Elsevier.

\bibitem[{Vaswani et~al.(2017)Vaswani, Shazeer, Parmar, Uszkoreit, Jones,
  Gomez, Kaiser, and Polosukhin}]{vaswani2017attention}
Ashish Vaswani, Noam Shazeer, Niki Parmar, Jakob Uszkoreit, Llion Jones,
  Aidan~N Gomez, {\L}ukasz Kaiser, and Illia Polosukhin. 2017.
\newblock Attention is all you need.
\newblock In \emph{Advances in Neural Information Processing Systems}, pages
  5998--6008.

\bibitem[{Wang et~al.(2017)Wang, Hamza, and Florian}]{wang2017bilateral}
Zhiguo Wang, Wael Hamza, and Radu Florian. 2017.
\newblock Bilateral multi-perspective matching for natural language sentences.
\newblock \emph{IJCAI}.

\bibitem[{Xia(2000)}]{xia2000part}
Fei Xia. 2000.
\newblock The part-of-speech tagging guidelines for the penn chinese treebank
  (3.0).
\newblock \emph{IRCS Technical Reports Series}, page~38.

\bibitem[{Xu et~al.(2004)Xu, Zens, and Ney}]{xu}
Jia Xu, Richard Zens, and Hermann Ney. 2004.
\newblock Do we need chinese word segmentation for statistical machine
  translation?
\newblock In \emph{Proceedings of the Third SIGHAN Workshop on Chinese Language
  Processing}.

\bibitem[{Yang et~al.(2017)Yang, Zhang, and Dong}]{yang2017neural}
Jie Yang, Yue Zhang, and Fei Dong. 2017.
\newblock Neural word segmentation with rich pretraining.
\newblock \emph{ACL}.

\bibitem[{Yang et~al.(2018)Yang, Zhang, and Liang}]{yang2018subword}
Jie Yang, Yue Zhang, and Shuailong Liang. 2018.
\newblock Subword encoding in lattice lstm for chinese word segmentation.
\newblock \emph{arXiv preprint arXiv:1810.12594}.

\bibitem[{Yao and Huang(2016)}]{yao2016bi}
Yushi Yao and Zheng Huang. 2016.
\newblock Bi-directional lstm recurrent neural network for chinese word
  segmentation.
\newblock In \emph{International Conference on Neural Information Processing},
  pages 345--353. Springer.

\bibitem[{Yin et~al.(2016)Yin, Wang, Li, Li, and Wang}]{yin2016multi}
Rongchao Yin, Quan Wang, Peng Li, Rui Li, and Bin Wang. 2016.
\newblock Multi-granularity chinese word embedding.
\newblock In \emph{Proceedings of the 2016 Conference on Empirical Methods in
  Natural Language Processing}, pages 981--986.

\bibitem[{Yu et~al.(2017)Yu, Jian, Xin, and Song}]{yu2017joint}
Jinxing Yu, Xun Jian, Hao Xin, and Yangqiu Song. 2017.
\newblock Joint embeddings of chinese words, characters, and fine-grained
  subcharacter components.
\newblock In \emph{Proceedings of the 2017 Conference on Empirical Methods in
  Natural Language Processing}, pages 286--291.

\bibitem[{Yu et~al.(2001)Yu, Lu, Zhu, Duan, Kang, Sun, Wang, Zhao, and
  Zhan}]{yu2001processing}
Shiwen Yu, Jianming Lu, Xuefeng Zhu, Huiming Duan, Shiyong Kang, Honglin Sun,
  Hui Wang, Qiang Zhao, and Weidong Zhan. 2001.
\newblock Processing norms of modern chinese corpus.
\newblock Technical report, Technical report.

\bibitem[{Zhang et~al.(2003)Zhang, Yu, Xiong, and Liu}]{zhang2003hhmm}
Hua-Ping Zhang, Hong-Kui Yu, De-Yi Xiong, and Qun Liu. 2003.
\newblock Hhmm-based chinese lexical analyzer ictclas.
\newblock In \emph{Proceedings of the second SIGHAN workshop on Chinese
  language processing-Volume 17}, pages 184--187. Association for Computational
  Linguistics.

\bibitem[{Zhang et~al.(2018)Zhang, Liu, Luan, Xu, and Sun}]{zhang2018prior}
Jiacheng Zhang, Yang Liu, Huanbo Luan, Jingfang Xu, and Maosong Sun. 2018.
\newblock Prior knowledge integration for neural machine translation using
  posterior regularization.
\newblock \emph{arXiv preprint arXiv:1811.01100}.

\bibitem[{Zhang et~al.(2017)Zhang, Fu, and Yu}]{zhang2017segmenting}
Meishan Zhang, Guohong Fu, and Nan Yu. 2017.
\newblock Segmenting chinese microtext: Joint informal-word detection and
  segmentation with neural networks.
\newblock In \emph{IJCAI}, pages 4228--4234.

\bibitem[{Zhang et~al.(2016)Zhang, Zhang, and Fu}]{zhang2016transition}
Meishan Zhang, Yue Zhang, and Guohong Fu. 2016.
\newblock Transition-based neural word segmentation.
\newblock In \emph{Proceedings of the 54th Annual Meeting of the Association
  for Computational Linguistics (Volume 1: Long Papers)}, volume~1, pages
  421--431.

\bibitem[{Zhang and LeCun(2017)}]{zhang2017encoding}
Xiang Zhang and Yann LeCun. 2017.
\newblock Which encoding is the best for text classification in chinese,
  english, japanese and korean?
\newblock \emph{arXiv preprint arXiv:1708.02657}.

\bibitem[{Zhao et~al.(2006)Zhao, Huang, and Li}]{zhao2006improved}
Hai Zhao, Chang-Ning Huang, and Mu~Li. 2006.
\newblock An improved chinese word segmentation system with conditional random
  field.
\newblock In \emph{Proceedings of the Fifth SIGHAN Workshop on Chinese Language
  Processing}, pages 162--165.

\bibitem[{Zhao et~al.(2013)Zhao, Utiyama, Sumita, and Lu}]{zhao2013empirical}
Hai Zhao, Masao Utiyama, Eiichiro Sumita, and Bao-Liang Lu. 2013.
\newblock An empirical study on word segmentation for chinese machine
  translation.
\newblock In \emph{International Conference on Intelligent Text Processing and
  Computational Linguistics}, pages 248--263. Springer.

\bibitem[{Zheng et~al.(2013)Zheng, Chen, and Xu}]{zheng2013deep}
Xiaoqing Zheng, Hanyang Chen, and Tianyu Xu. 2013.
\newblock Deep learning for chinese word segmentation and pos tagging.
\newblock In \emph{Proceedings of the 2013 Conference on Empirical Methods in
  Natural Language Processing}, pages 647--657.

\bibitem[{Zhou et~al.(2017)Zhou, Yu, Zhang, Huang, DAI, and
  Chen}]{zhou2017word}
Hao Zhou, Zhenting Yu, Yue Zhang, Shujian Huang, XIN-YU DAI, and Jiajun Chen.
  2017.
\newblock Word-context character embeddings for chinese word segmentation.
\newblock In \emph{Proceedings of the 2017 Conference on Empirical Methods in
  Natural Language Processing}, pages 760--766.

\bibitem[{Zhuang et~al.(2010)Zhuang, Luo, Xiong, Xiong, He, and
  Shi}]{zhuang2010cross}
Fuzhen Zhuang, Ping Luo, Hui Xiong, Yuhong Xiong, Qing He, and Zhongzhi Shi.
  2010.
\newblock Cross-domain learning from multiple sources: A consensus
  regularization perspective.
\newblock \emph{IEEE Transactions on Knowledge and Data Engineering},
  22(12):1664--1678.

\end{thebibliography}
\bibliographystyle{acl_natbib}

\end{document}